\documentclass[runningheads, envcountsame, a4paper]{llncs}
\usepackage{bibnames}

\usepackage{amsmath,amssymb,amsfonts}
\usepackage{algorithmic}
\usepackage{mathtools}

\usepackage{graphicx}
\usepackage{textcomp}
\usepackage{booktabs}
\usepackage{url}
\usepackage{paralist}
\usepackage{todonotes}
\usepackage{hyperref}
\usepackage{array}
\usepackage{makecell}
\usepackage[export]{adjustbox}

\usepackage{epstopdf}

\usepackage{multirow}

\definecolor{darkred}{rgb}{.8,0,0}
\definecolor{darkgreen}{rgb}{0,.5,0}

\usepackage{microtype}

\usepackage{xcolor}
\def\BibTeX{{\rm B\kern-.05em{\sc i\kern-.025em b}\kern-.08em
    T\kern-.1667em\lower.7ex\hbox{E}\kern-.125emX}}

\begin{document}
\title{A Comparative Study of Transformers on \\ Word Sense Disambiguation}

\author{Avi Chawla\inst{1} \and
Nidhi Mulay\inst{1} \and
Vikas Bishnoi\inst{1} \and
Gaurav Dhama\inst{1} \and \\
Anil Kumar Singh \inst{2}}
\authorrunning{A. Chawla et al.}

\institute{Mastercard AI, Gurgaon, India \\
\email{$\{$avi.chawla, nidhi.mulay, vikas.bishnoi, gaurav.dhama$\}$@mastercard.com} \and
Indian Institute of Technology (BHU), Varanasi, India\\
\email{aksingh.cse@iitbhu.ac.in}}

\maketitle              
\begin{abstract}
Recent years of research in Natural Language Processing (NLP) have witnessed dramatic growth in training large models for generating context-aware language representations. In this regard, numerous NLP systems have leveraged the power of neural network-based architectures to incorporate sense information in embeddings, resulting in Contextualized Word Embeddings (CWEs). Despite this progress, the NLP community has not witnessed any significant work performing a comparative study on the contextualization power of such architectures. This paper presents a comparative study and an extensive analysis of nine widely adopted Transformer models. These models are BERT, CTRL, DistilBERT, OpenAI-GPT, OpenAI-GPT2, Transformer-XL, XLNet, ELECTRA, and ALBERT. We evaluate their contextualization power using two lexical sample Word Sense Disambiguation (WSD) tasks, SensEval-2 and SensEval-3. We adopt a simple yet effective approach to WSD that uses a \emph{k}-Nearest Neighbor (kNN) classification on CWEs. Experimental results show that the proposed techniques also achieve superior results over the current state-of-the-art on both the WSD tasks.

\end{abstract}
\section{Introduction}
Developing powerful language representations technique has been a key area of research in Natural Language Processing (NLP). The employment of effective representational models has also been an essential contributor in improving the performance of many NLP systems. Word vectors or embeddings are fixed-length vectors that are proficient in capturing the semantic properties of words. Emerging from a simple Neural Network-based Word2Vec model and recently transitioning to Contextualised Word Embeddings (CWEs), the advancements have consistently brought a revolution to every NLP sub-domain. The introduction of naive Word2Vec model not only brought an unprecedented increase in the performance of a wide variety of downstream tasks such as Machine Translation, Sentiment Analysis, and Question Answering, but it also laid the foundation for a majority of Natural Language Understanding (NLU) architectures that we use today.

Recent attempts in NLU research have fundamentally focused on generating context-aware word representations, i.e., embeddings that take into account the polysemous nature of words. Polysemy refers to the changes in the meaning of a word when the context around it changes. One related task in NLP is Word Sense Disambiguation (WSD) which deals with the automatic recognition of the correct sense of a word appearing in a specific context. WSD is an essential component of any NLP system as it helps in generating better semantic representations of words.

\noindent \textbf{Contribution}: The Transformer architectures implemented in the HuggingFace  framework \cite{Wolf2019HuggingFacesTS} implicitly provide a model for WSD. We test the performance of nine such pre-trained models on WSD and extensively analyse each one of them. These models are BERT \cite{devlin2018bert}, OpenAI-GPT \cite{radford2018gpt}, OpenAI-GPT2 \cite{radford2019language}, CTRL \cite{keskarCTRL2019}, DistilBERT \cite{sanh2019distilbert}, Transformer-XL \cite{DBLP:journals/corr/abs-1901-02860}, XLNet \cite{DBLP:journals/corr/abs-1906-08237}, ELECTRA \cite{clark2020electra} and ALBERT \cite{lan2019albert}.
This comprehensive study helps us in comparing the ability of different transformer models in incorporating polysemy in embeddings, i.e., their power of segregating various senses of a word in the word-vector space. Through our experiments, we also report a new state-of-the-art on both the lexical sample WSD datasets we experimented on, i.e., SensEval-2 and SensEval-3.

\noindent \textbf{Note}: Although the prime use of the CTRL, OpenAI-GPT, and OpenAI-GPT2 model is Natural Language Generation (NLG), we still include them in our comparative study. We do this to determine the extent to which these models consider polysemy while carrying out NLG as their primary objective.

\section{Related Work}


Word Sense Disambiguation (WSD) is an old and common problem in NLP. In the early days of Artificial Intelligence, WSD was conceived as a fundamental task of Machine Translation \cite{weaver.1949}. Since then, advancements in NLP have led to the development of a variety of WSD systems. Recent attempts in this respect have tried to tackle the problem by introducing the concept of sense embeddings. For instance, \cite{bartunovetal:2016} induced sense embeddings using a pre-training based approach. \cite{pelevina:2016:RepL4NLP} proposed methods that focus on generating sense embeddings using pre-trained word embeddings such as Glove vectors \cite{pennington2014glove}. 
\cite{DBLP:journals/corr/TraskML15} proposed `Sense2Vec', which utilized the part-of-speech and named entity tag information to distinguish between different meanings of a word. 
An extensive survey on further ideas and research on sense representations of words is given by \cite{DBLP:journals/corr/abs-1805-04032}.

Most of the recent approaches have leveraged the power of Deep Learning to build WSD systems. \cite{bosc-vincent-2018-auto} proposed an auto-encoder-based approach that goes from the target word embedding back to the word definition. The method proposed by \cite{yuan.2016} revolved around the computation of sentence context vector for ambiguous words. They adopted a \emph{k}-Nearest Neighbor (kNN) \cite{cover.1967} based approach for classification of ambiguous words. In contrast to all the approaches described above, \cite{Wiedemann2019DoesBM} proposed a simple yet effective approach for the classification of ambiguous words. Instead of using any pre-trained embeddings like the glove embeddings, they used the BERT \cite{devlin2018bert} to obtain Contextualised Word Embeddings (CWEs). For prediction, they used a kNN based approach. The use of BERT also achieved new state-of-the-art results over previously proposed approaches.

\section{Datasets}

In our experiments, we use two widely-adopted lexical sample corpora available for WSD, SensEval-2 and SensEval-3. Both come with a train and test set to train and evaluate a WSD model. The words in these datasets are annotated with the sense identifiers defined in WordNet 3.0. A brief overview of both datasets is shown in Table 1. To evaluate the performance of a WSD model, we refer to the testing scripts from the comprehensive framework of \cite{raganato.2017}\footnote{\href{https://github.com/getalp/UFSAC}{https://github.com/getalp/UFSAC}.}. 

\begin{table*}[t]
\caption{An overview of the datasets used for the study of nine Transformer Models. Average Sentence Length has been rounded off to the nearest integer.}
\resizebox{\textwidth}{!}
{%
\small
\begin{tabular}{@{}ccccccccc@{}}
  \toprule
   \textbf{Dataset} & \textbf{\thead{No. of \\ Sentences}} & \textbf{\thead{Avg. Sentence \\ Length}} & \textbf{\thead{No. of Distinct \\ Sense Identifiers}} & \textbf{\thead{No. of Sense \\ Embeddings}}  & \textbf{\thead{Distinct \\ Words}} & \textbf{Nouns}  & \textbf{Adjectives} & \textbf{Verbs}  \\
  \midrule
  SensEval-3 Train & 7860 & 30 & 285 & 9280 & 172 & 3632 & 308 & 3879 \\
  SensEval-3 Test  & 3944 & 30 & 260 & 4520 & 168 & 1777 & 153 & 1999  \\
  SensEval-2 Train & 8611 & 29 & 783 & 8742 & 187 & 3492 & 1400 & 2559 \\
  SensEval-2 Test  & 4328 & 29 & 620 & 4385 & 184 & 1737 & 702 & 1800  \\
 
  \bottomrule
\end{tabular}%
}

\end{table*}

\section{Experiments}
For our experimentation, we take inspiration from a simple yet effective kNN based approach on CWEs to WSD proposed by \cite{Wiedemann2019DoesBM}.
This approach uses a cosine similarity-based distance metric for the classification of ambiguous words in the test data. In a nutshell, we obtain the CWEs of all the ambiguous words in the training data by providing their respective contexts to one of the nine contextualization approaches. While classifying an ambiguous word in a test sentence, a kNN classification approach is used, with cosine similarity between the CWE of the ambiguous word and all its instances observed during training as the similarity metric. 
Such an experiment is carried out for six different values of the hyper-parameter $k \in \lbrace 1, 3, 5, 7, 10, 11 \rbrace$ in the kNN classifier. An ambiguous word is classified to the sense with the maximum number of nearest neighbors in the ``\emph{k}'' nearest neighbors. 

We propose few additions to existing approach to improve the overall performance. The first improvement lies in the way data is collected. \cite{Wiedemann2019DoesBM} used the lemma of every word in a sentence to obtain sentences from the dataset. This, in some cases, generated inappropriate sentences such as: 
\emph{``Nor be this feeling only provoked by the sight or the thought of art, he write."} instead of \emph{``Nor is this feeling only provoked by the sight or the thought of art, he wrote."}. Another sentence collected by their method and our method is \emph{``The art\_critic critic be thus bind to consider with care what standard of comparison should be use."} and \emph{``The art critic is thus bound to consider with care what standards of comparison should be used."} respectively. The sentences collected by their method lack a proper grammatical sense and structure. We improve this by collecting the lemma only for ambiguous words and surface form for every other word in the sentence. 

Our second improvement is an empirical finding. While obtaining the CWEs from BERT, they treated the concatenation of the output of the last four layers of BERT as the word embeddings. Instead, we used only the final layer of BERT to obtain the embedding of a word.

    

\section{Experimental Results}

To study and analyze each of the transformer models in detail, we conduct three rounds of experiments. In the first round, we carry out the task of WSD on nine pre-trained Transformer architectures using the kNN approach described above and compare their performance on two WSD Lexical Sample tasks. Further, to visualize each model's power to separate different senses of a word in their embedding space, we draw the t-SNE plots for the CWEs generated by the transformer models. Lastly, we provide a qualitative analysis by examining the correct predictions and the wrong predictions made by each of our WSD models. 

\subsection{Contextualized Embeddings}

To compare the models based on their contextualization power, we perform the task of WSD using the language representation provided by them. Table 2 lists the results obtained by each of these models for $k \in \lbrace 1, 3, 5, 7, 10, 11 \rbrace$. The BERT model achieved a new state-of-the-art on the SensEval-2 and SensEval-3 tasks \cite{Wiedemann2019DoesBM}. The modification also facilitates the DistilBERT model in beating the current state-of-the-art on SensEval-3 dataset. Also, it becomes evident from the results obtained by the ALBERT and the DistilBERT model that they highly resemble BERT's architecture. Through our observations, we also state that the employment of DistilBERT and ALBERT in place of BERT could take off a major overhead of the training time without incurring a significant loss in the performance. 

An unexpected drop in performance is observed for the XLNet and Transformer-XL model compared to the other well-performing models. Though both the models are effective on various NLP tasks using their powerful recurrence-based Transformer architectures, we notice that the model still underperforms. 

Coming towards the end, the three NLG models — OpenAI-GPT, OpenAI-GPT2, and CTRL also performed poorly. CTRL and OpenAI-GPT2 performed slightly better than the Most Frequent Sense (MFS) baseline on SensEval-3 dataset. In addition to this, they even failed to beat the MFS baseline on the SensEval-2 dataset, demonstrating that they are ineffective in capturing polysemy.
\begin{table}[t]

\caption{Results(F1\%) of all the Transformer models for different values of \emph{k} on the \emph{k}-Nearest Neighbor classification approach vs. the Most Frequent Sense (MFS) baseline and current state-of-the-art results. The Best results for each model are underlined, and the best result on a particular dataset is bold. The previous state-of-the-art is in italics.}
\resizebox{\textwidth}{!}
{%

\begin{tabular}{@{}p{0.02\textwidth}|c|c|c|c|c|c||c|c|c|c|c|c@{}}
\toprule
\multicolumn{1}{c|}{\textbf{Model}} & \multicolumn{6}{c||}{\textbf{SensEval-2}} & \multicolumn{6}{c}{\textbf{SensEval-3}} \\
& {\small k=1} & {\small k=3} & {\small k=5} & {\small k=7} & {\small k=10} & {\small k=11} & {\small k=1} & {\small k=3} & {\small k=5} & {\small k=7} & {\small k=10} & {\small k=11} \\
\midrule
\multicolumn{1}{c|}{BERT} & 76.02 & 76.78 & 76.62 & 76.62 & 76.76 & \underline{\textbf{76.81}} & 79.40 & 80.31 & 80.49 & \underline{\textbf{80.96}} & 80.75 & 80.72 \\
\multicolumn{1}{c|}{DistilBERT} & 74.81 & \underline{75.64} & 75.36 & 75.43 & 75.41 & 75.43 & 78.62 & 79.71 & 80.05 & 80.15 & \underline{80.23} & 80.07 \\
\multicolumn{1}{c|}{ALBERT} & 74.84 & 75.33 & \underline{75.43} & 74.98 & 75.07 & 75.07 & 77.94 & 78.93 & 79.44 & 79.60 & \underline{79.71} & 79.57 \\
\multicolumn{1}{c|}{XLNet}  & 64.74 & 66.24 & \underline{66.48} & 66.45 & 66.38 & 66.45 & 69.97 & 70.64 & 71.50 & \underline{71.78} & 71.42 & 71.42 \\
\multicolumn{1}{c|}{ELECTRA} & 65.98 & 65.88 & 65.98 & \underline{66.10} & 66.07 & 65.95 & 69.45 & 70.10 & 70.82 & \underline{71.14} & 71.11 & 71.01 \\
\multicolumn{1}{c|}{GPT}  & 59.80 & 60.84 & 61.29 & 61.24 & 61.15 & \underline{61.54} & 65.63 & 67.65 & 68.51 & 69.29 & 69.58 & \underline{69.60} \\
\multicolumn{1}{c|}{Trans-XL} & 53.36 & 54.35 & 55.01 & \underline{55.18} & 55.01 & 54.45 & 62.07 & 62.82 & 63.32 & \underline{63.99} & 63.50 & 63.50 \\
\multicolumn{1}{c|}{CTRL} & 52.39 & 53.64 & 54.28 & 54.45 & 54.49 & \underline{54.82} & 58.09 & 60.38 & 60.92 & 61.50 & \underline{61.78} & 61.63 \\
\multicolumn{1}{c|}{GPT2}  & 50.96 & 53.57 & \underline{53.88} & \underline{53.88} & 53.86 & 53.80 & 57.03 & 59.83 & 60.92  & 61.21 & \underline{61.29} & 61.19 \\
\midrule
\multicolumn{1}{c|}{MFS} & \multicolumn{6}{c||}{54.79} & \multicolumn{6}{c}{58.95} \\
\multicolumn{1}{c|}{kNN \cite{Wiedemann2019DoesBM}} & \multicolumn{6}{c||}{\emph{76.52}} & \multicolumn{6}{c}{\emph{80.12}} \\
\bottomrule
\end{tabular}%
}

\end{table}


\subsection{Sense-space analysis using t-SNE plots}

To understand and interpret a model's power to segregate different senses of a word in the embedding space, we draw the t-SNE plots of the embeddings obtained for the word `bank' from the training data of SensEval-3 for each of the nine models. Figure 2 represents the t-SNE plots thus obtained. Sub-figure 2.(j) represents the interpretable meanings of the senses represented in the t-SNE plots along with their respective frequencies in the SensEval-3 training corpus. We exclude any sense with a frequency of less than three from the t-SNE plot for clarity. It is evident from the t-SNE plot of OpenAI-GPT2 that it hardly distinguishes between different senses, and we see this as a possible reason why its accuracy is very close to the MFS baseline. As all the sense embeddings are in the vicinity of each other, the model hardly learns any decision boundary for sense classification. Therefore, the approach performs slightly better than the MFS baseline. We can draw a similar conclusion by observing the t-SNE plots of CTRL and Transformer-XL, implying that the NLG objective of OpenAI-GPT2 and CTRL hardly takes polysemy into account.

\begin{figure*}
  \centering
  \begin{tabular}{ c @{\hspace{5pt}} c @{\hspace{5pt}} c @{\hspace{5pt}} c @{\hspace{5pt}} c}
    \fbox{\includegraphics[width=0.17\textwidth, clip, trim={70 100 150     140}]{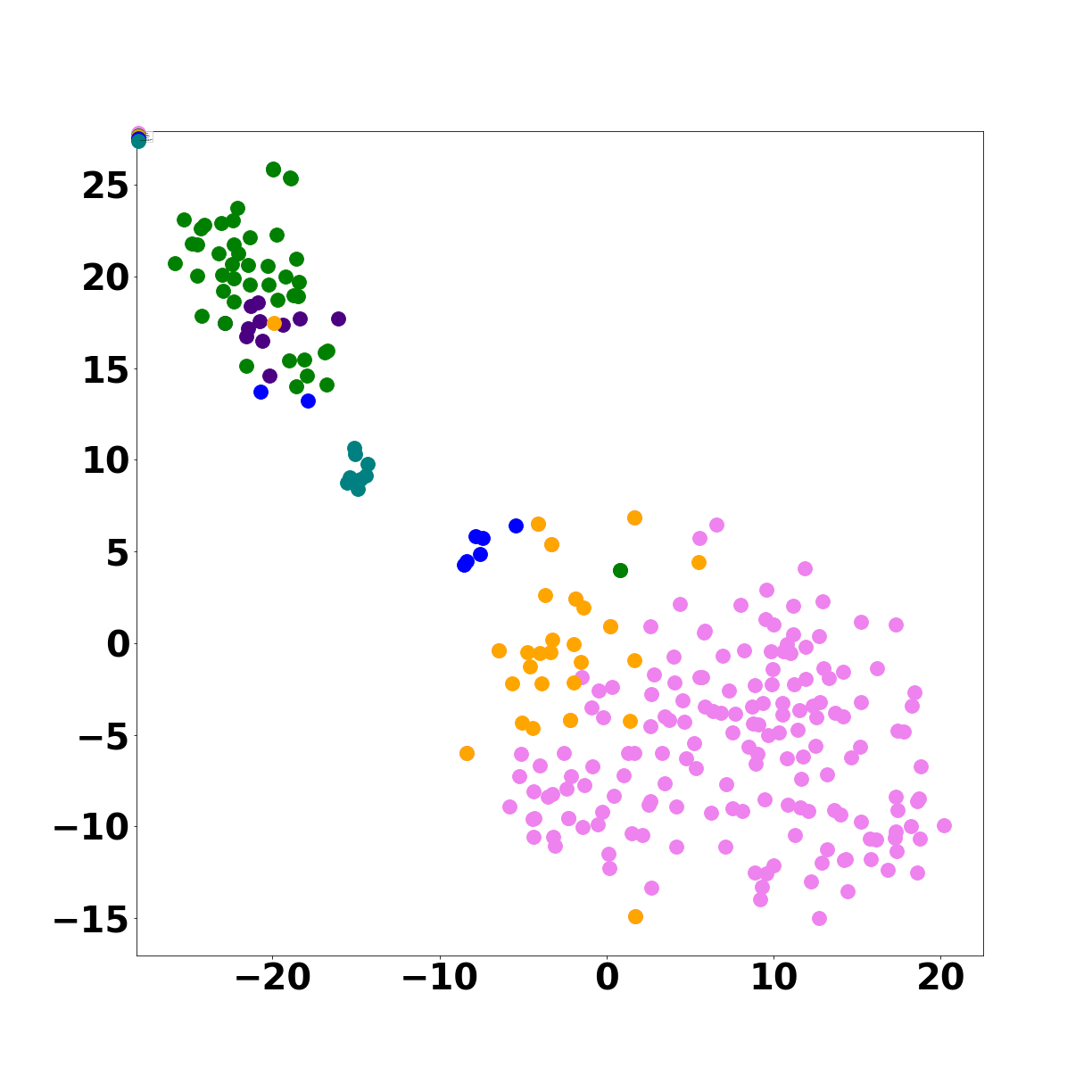}} &
     \fbox{\includegraphics[width=0.17\textwidth, clip, trim={70 100 150     140}]{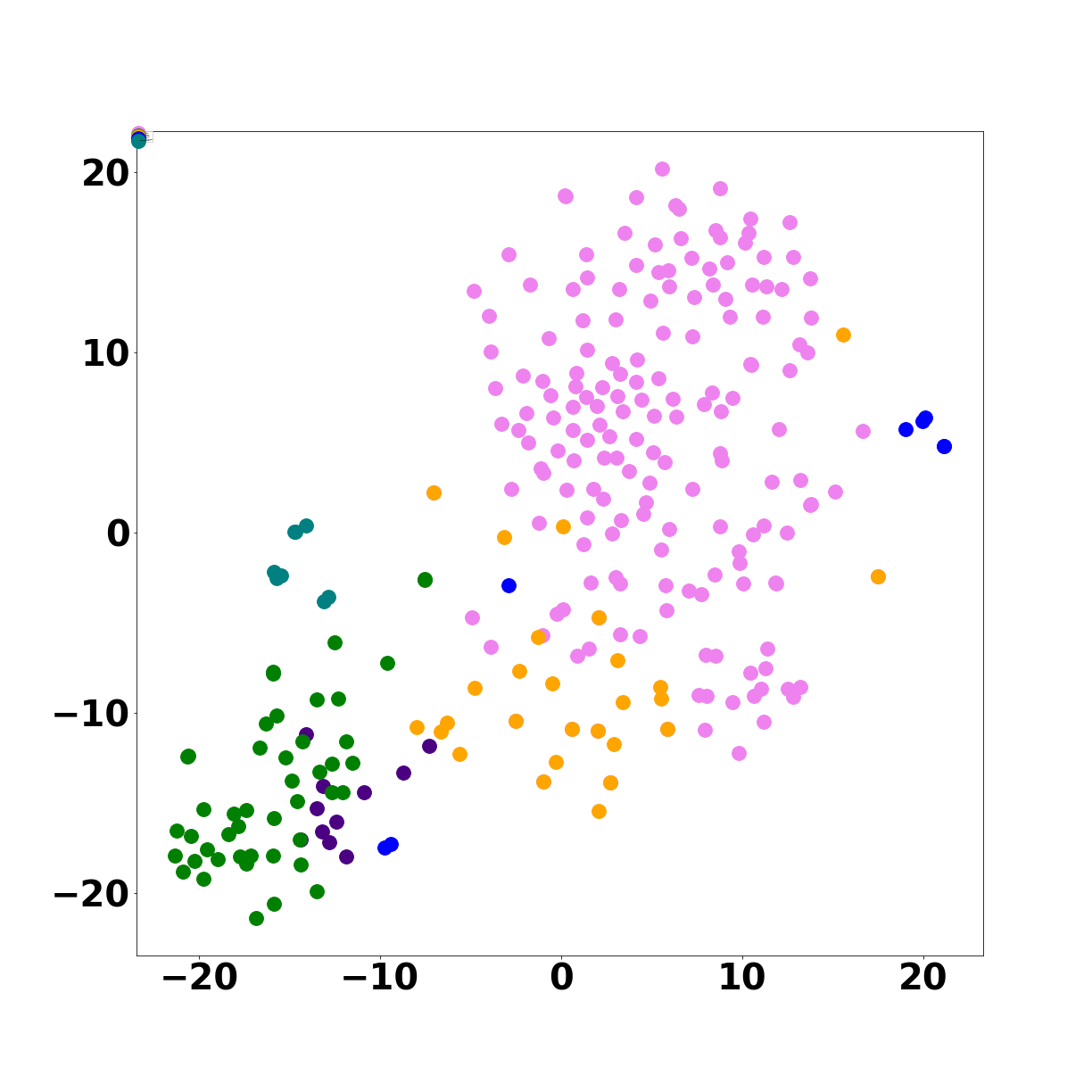}} &
      \fbox{\includegraphics[width=0.17\textwidth, clip, trim={70 100 150     140}]{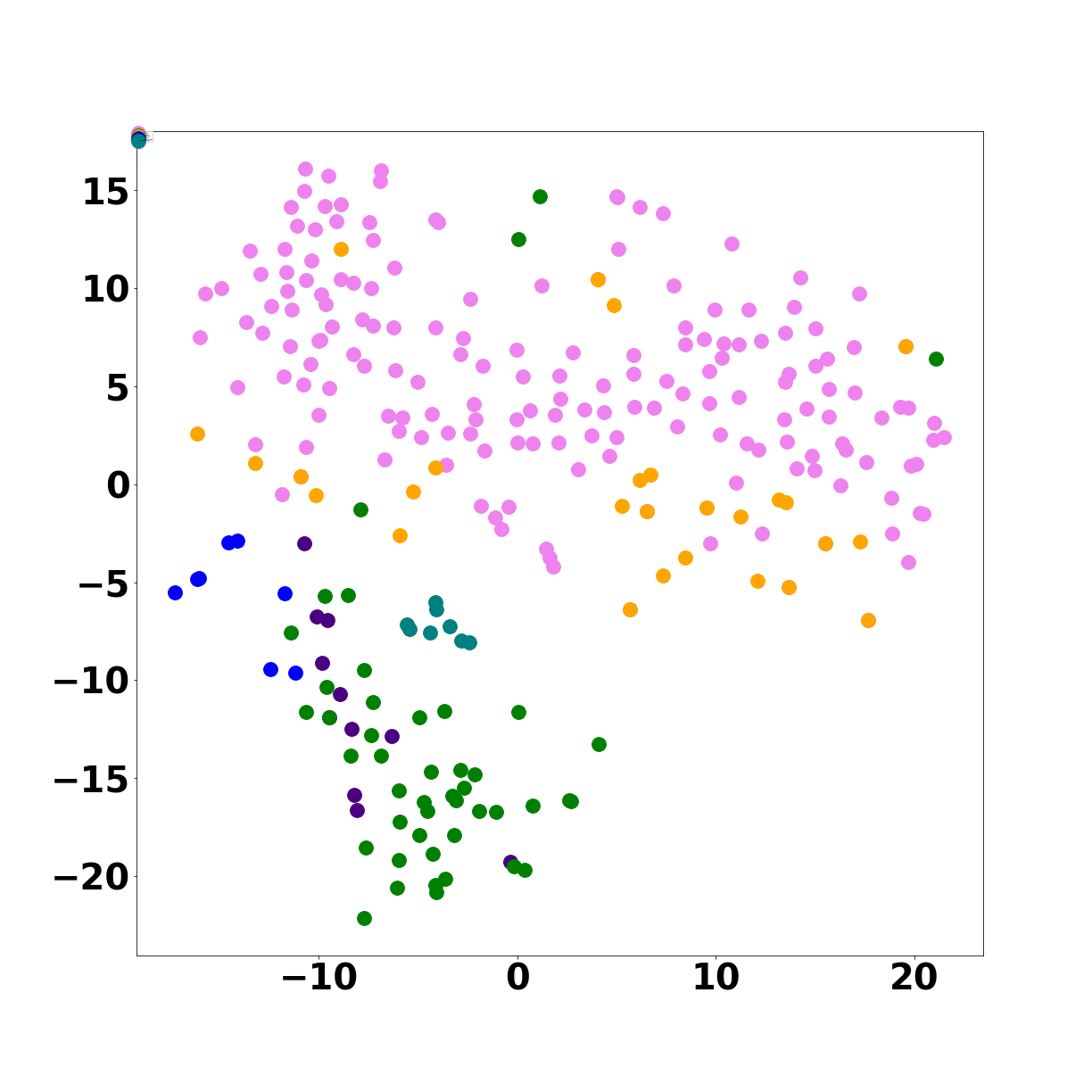}} &
      \fbox{\includegraphics[width=0.17\textwidth, clip, trim={70 100 150     140}]{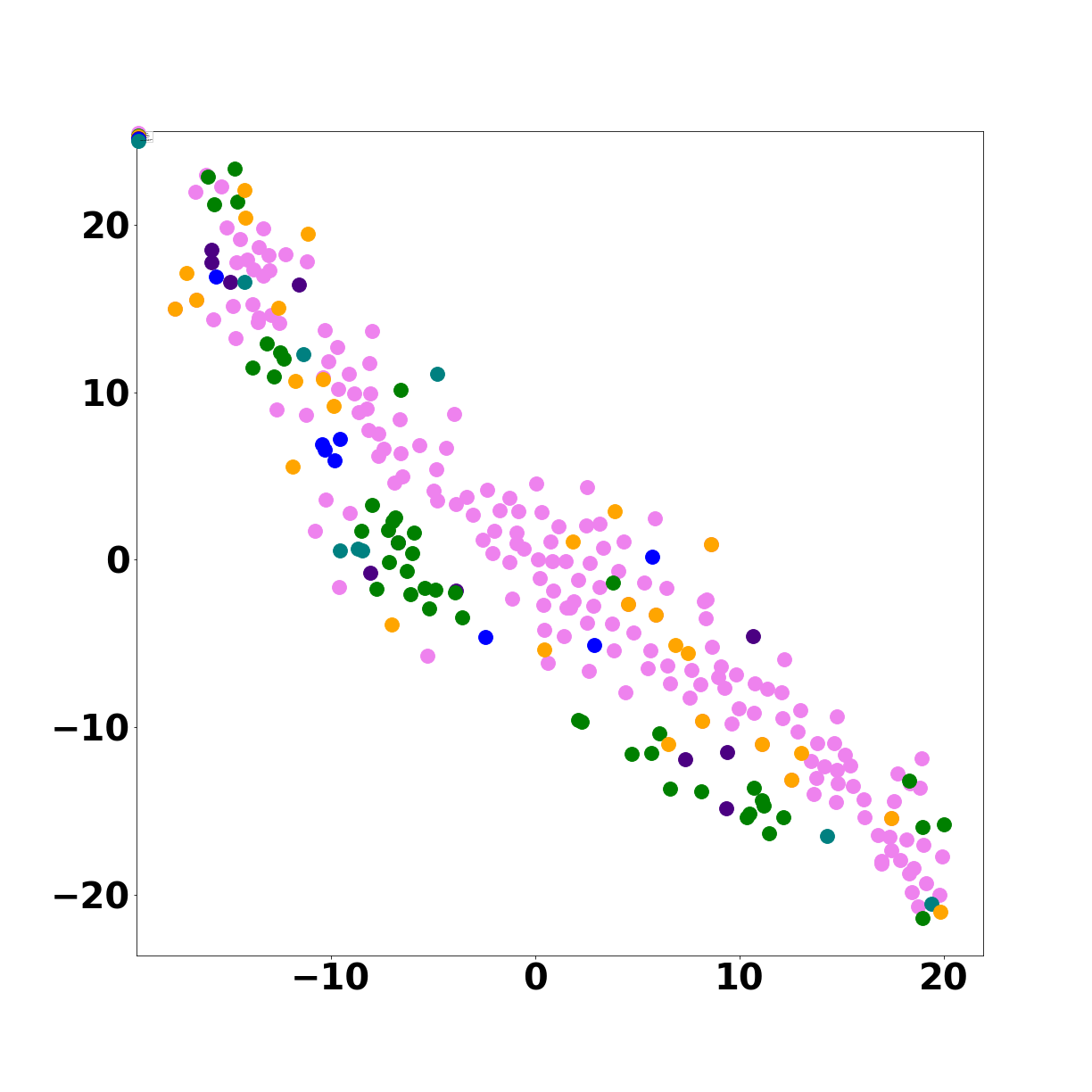}} &
      \fbox{\includegraphics[width=0.17\textwidth, clip, trim={70 100 150     140}]{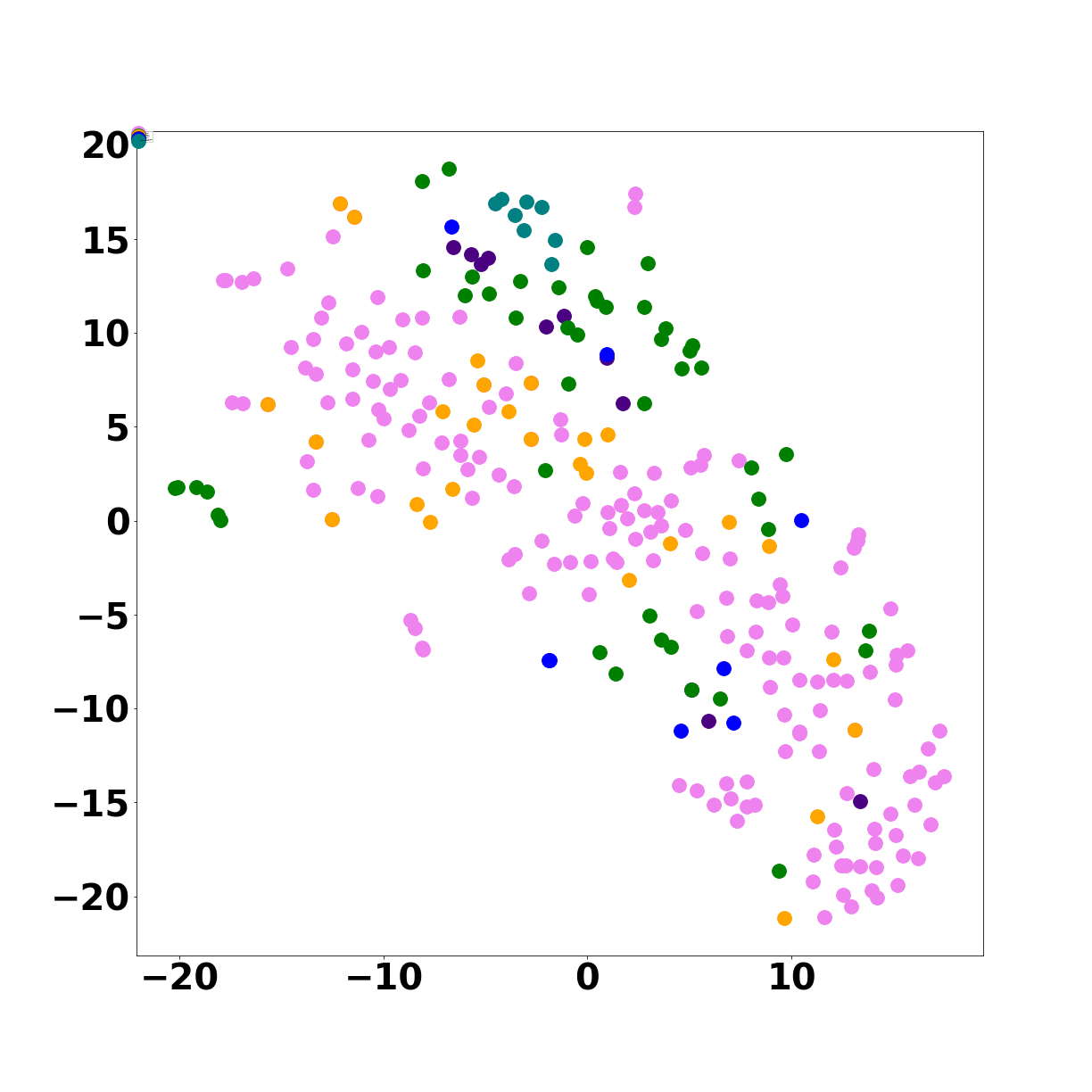}} \\
    \small (a) BERT &
      \small (b) DistilBERT &
      \small (c) ALBERT &
      \small (d) XLNet &
      \small (e) ELECTRA \\
      \end{tabular}
     \begin{tabular}{ c @{\hspace{5pt}} c @{\hspace{5pt}} c @{\hspace{5pt}} c @{\hspace{5pt}} c}
    \fbox{\includegraphics[width=0.17\textwidth, clip, trim={70 100 150     140}]{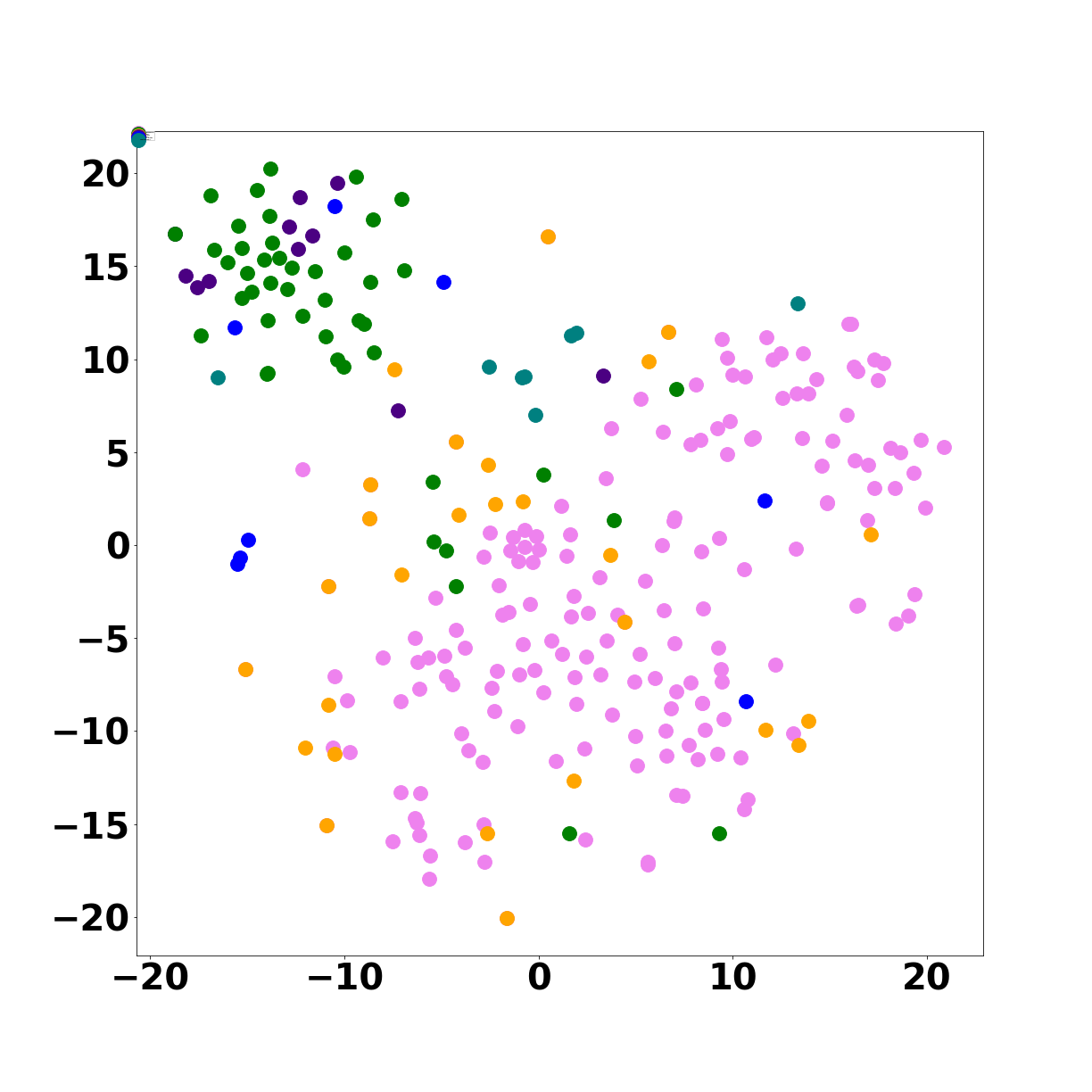}} &
     \fbox{\includegraphics[width=0.17\textwidth, clip, trim={70 100 150     140}]{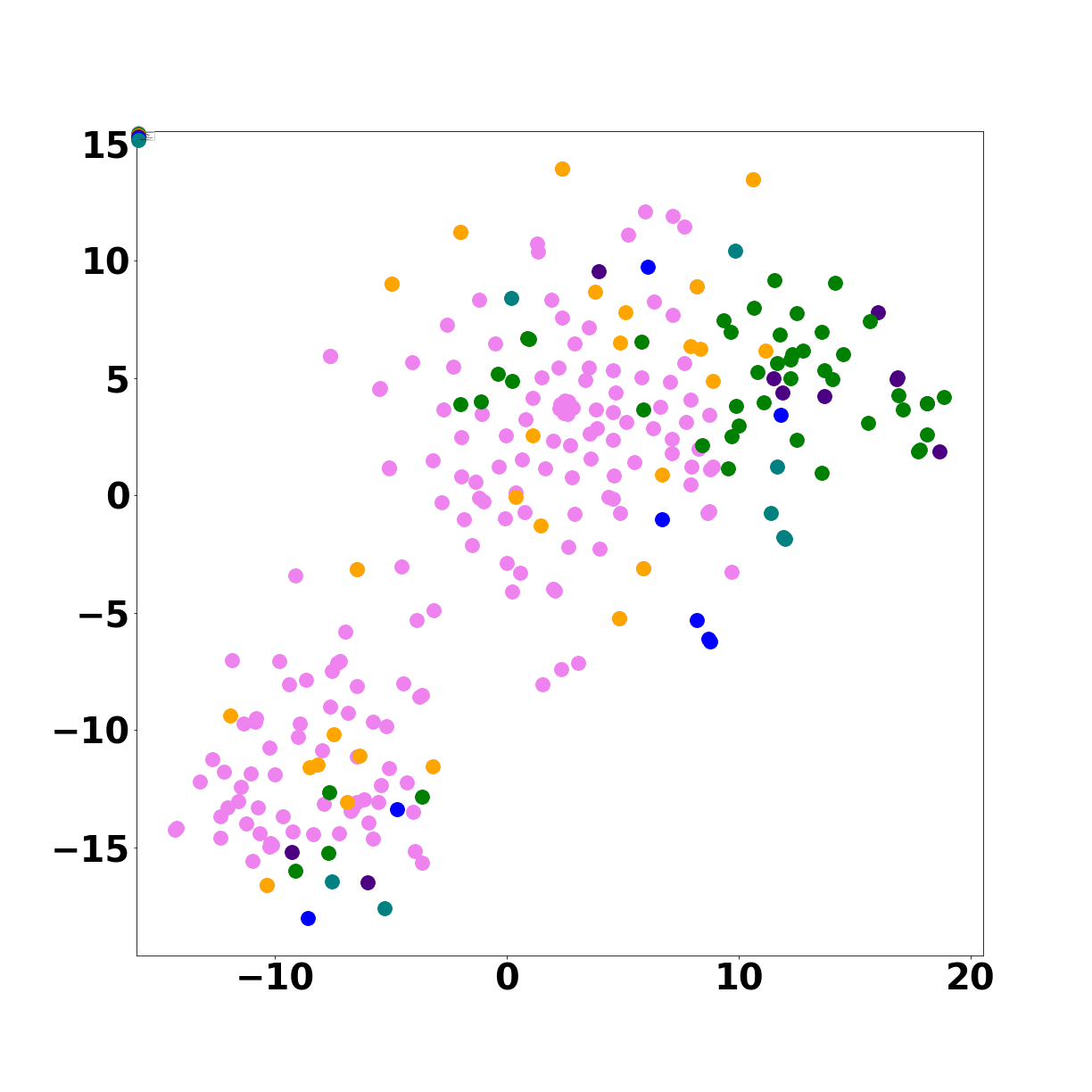}} &
      \fbox{\includegraphics[width=0.17\textwidth, clip, trim={70 100 150     140}]{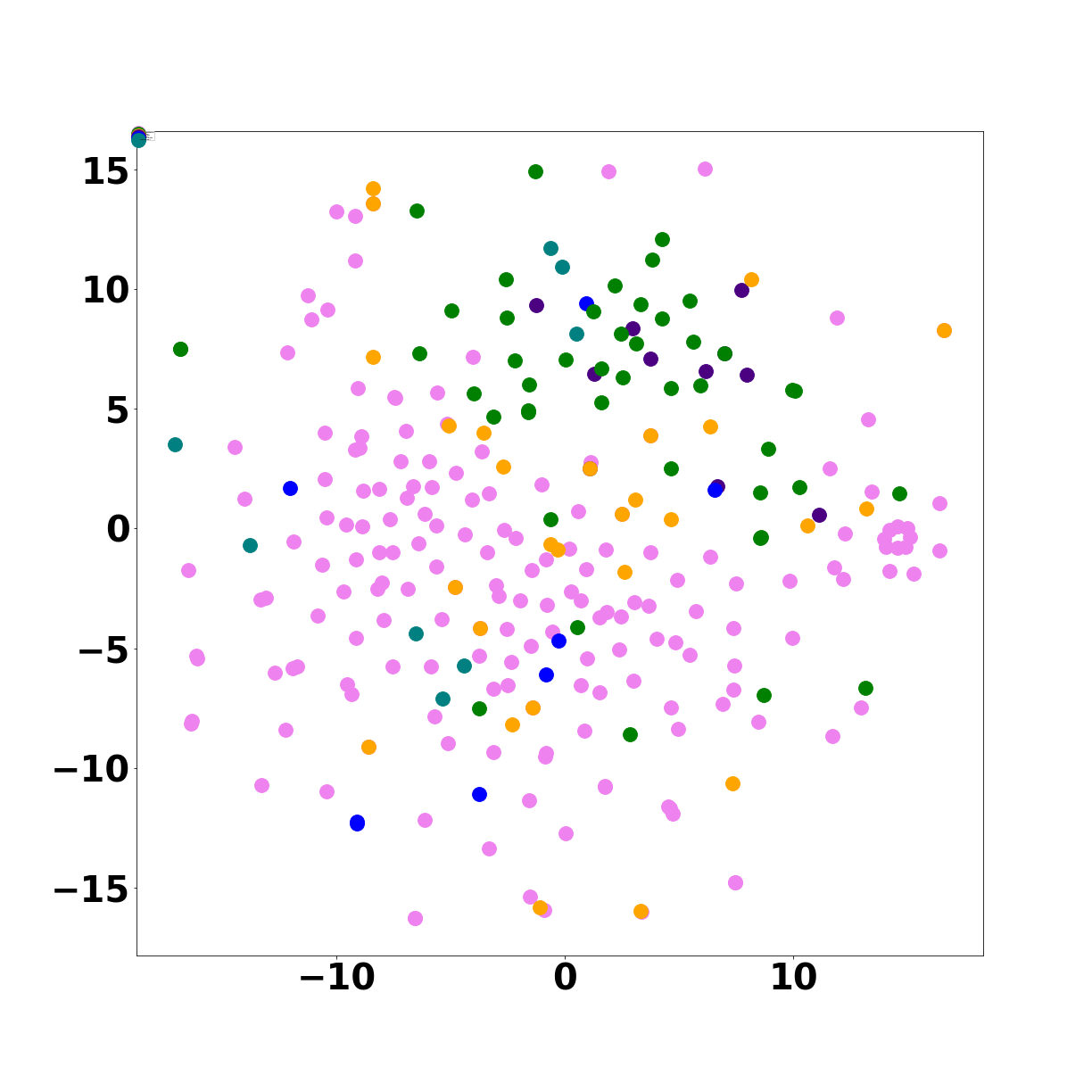}} &
      \fbox{\includegraphics[width=0.17\textwidth, clip, trim={70 100 150     140}]{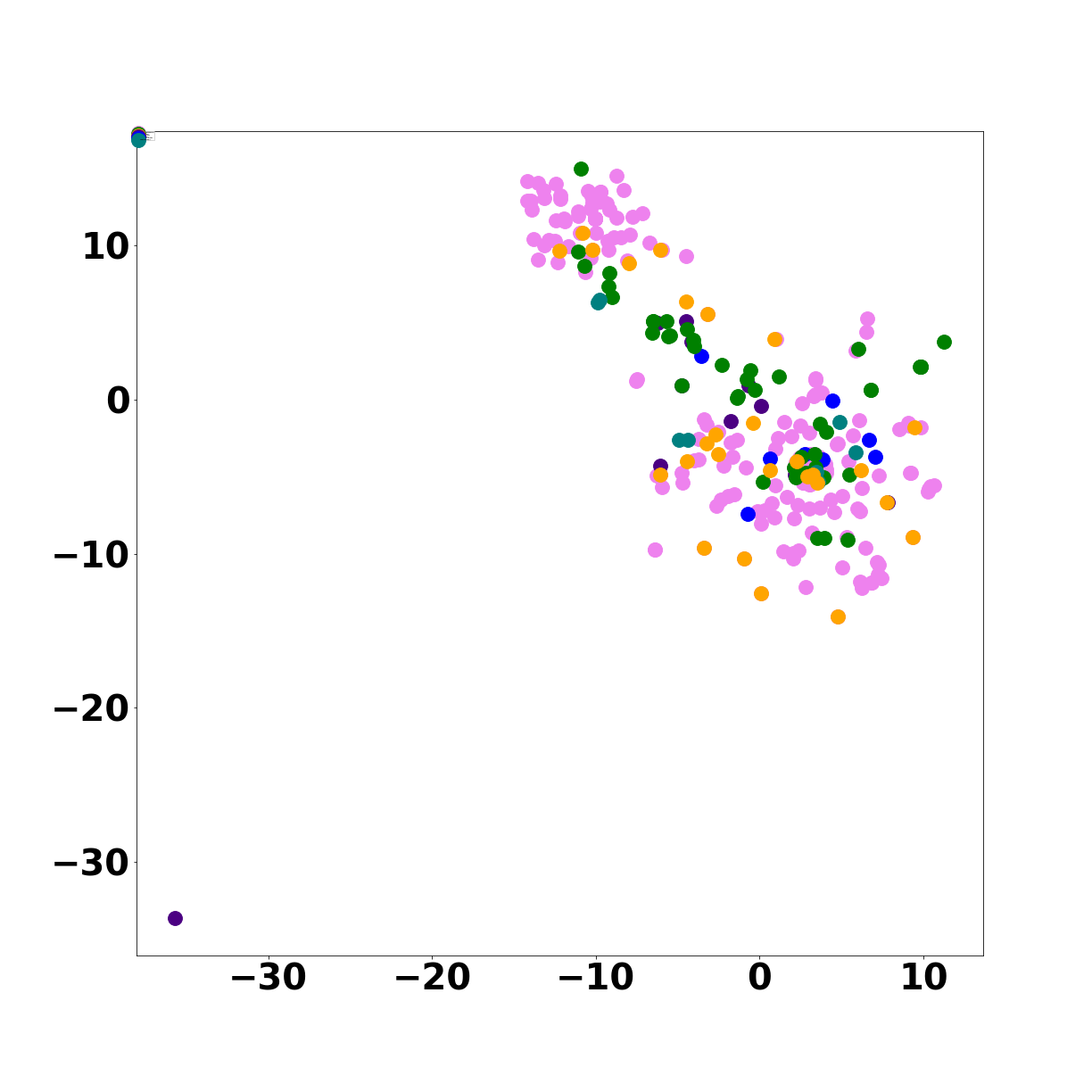}} &
      \fbox{\includegraphics[width=0.17\textwidth,
             height =2.030cm, clip, clip, trim={120 0 0 0}]{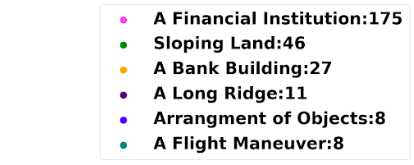}} \\
    \small (f) GPT&
      \small (g) Trans-XL&
      \small (h) CTRL &
      \small (i) GPT2&
      \small (j) Sense Labels\\
      
  \end{tabular}

  \medskip

  \caption{t-SNE plots of different senses of `bank' and their contextualized embeddings. The legend(shown separately in Sub-figure (j)) shows a short description of the respective WordNet sense and the frequency of occurrence in the training data. We used the SensEval-3 training dataset for obtaining these plots.}
\end{figure*}

On the other hand, plots obtained for the OpenAI-GPT, ELECTRA, and XLNet models depict that these models capture polysemy relatively better than the NLG models. They do stress a little on making a distinction between different senses of a word. Lastly, models that performed the best among the nine models we experimented on are BERT, DistilBERT, and ALBERT. These models possess exceptional proficiency in identifying polysemy, which is evident from their t-SNE plots as well as their accuracy on both the datasets.

\subsection{Additional Experiments}

Part-of-Speech information of a word has been regarded as a crucial influencer in determining its possible sense. \cite{Wiedemann2019DoesBM} proposed a POS-sensitive approach to WSD for the determination of the sense of an ambiguous word. Their experiments resulted in an accuracy lift of approximately 2-3 F1 on the SemEval datasets. Still, this approach did not prove to be beneficial for models trained on SensEval-2 and SensEval-3 datasets. This was because each word in these datasets is annotated with only one POS. 
Aligning our analysis on similar lines, as a final set of error analyses in our comparative study, we attempt to understand each model's behavior to different POS tags. We estimate the percentage of correct classifications made by each model for Nouns, Verbs, and Adjectives in the two datasets. This is presented in Table 4 for $\emph{k}=1$. 

For SensEval-2 dataset, we observe that each model was able to classify both Nouns and Adjectives correctly to a considerable extent. But, for Verbs, a difference of approximately 15-20\% was observed from that of Nouns and Adjectives. A similar drop in classification accuracy was observed in SensEval-3 for Adjectives. Each model classified Nouns and Verbs in this dataset to a reasonable extent but underperformed during the classification of Adjectives. 


\begin{table*}
\caption{The percentage of correct classifications made by the models for Nouns, Verbs and Adjectives on SensEval-2 Test and SensEval-3 Test data. The values are for \emph{k}=1. The best results on a particular POS tag have been marked in bold.}
\centering
{%
\begin{tabular}{@{}p{0.10\textwidth}|p{0.10\textwidth}p{0.10\textwidth}p{0.10\textwidth}|p{0.10\textwidth}p{0.10\textwidth}p{0.10\textwidth}@{}}
\toprule
\multicolumn{1}{c|}{\textbf{Model}} & \multicolumn{3}{c|}{\textbf{SensEval-2}} & \multicolumn{3}{c}{\textbf{SensEval-3}} \\
 & ~\hfill~{\small Nouns}~\hfill~ & ~\hfill~{\small Verbs}~\hfill~ & ~\hfill~{\small Adj}~\hfill~ & ~\hfill~{\small Nouns}~\hfill~ & ~\hfill~{\small Verbs}~\hfill~ & ~\hfill~{\small Adj}~\hfill~ \\
\midrule
\multicolumn{1}{c|}{BERT}  & ~\hfill~\textbf{81.64}~\hfill~ & ~\hfill~\textbf{67.22}~\hfill~ & ~\hfill~\textbf{81.62}~\hfill~ & ~\hfill~\textbf{78.17}~\hfill~ & ~\hfill~82.33~\hfill~ & ~\hfill~\textbf{56.86}~\hfill~ \\
\multicolumn{1}{c|}{DistilBERT}  & ~\hfill~81.00~\hfill~ & ~\hfill~65.61~\hfill~ & ~\hfill~80.06~\hfill~ & ~\hfill~76.14~\hfill~ & ~\hfill~\textbf{82.86}~\hfill~ & ~\hfill~54.25~\hfill~ \\
\multicolumn{1}{c|}{ALBERT}  & ~\hfill~82.38~\hfill~ & ~\hfill~64.33~\hfill~ & ~\hfill~80.06~\hfill~ & ~\hfill~75.63~\hfill~ & ~\hfill~81.87~\hfill~ & ~\hfill~55.56~\hfill~ \\
\multicolumn{1}{c|}{XLNet} & ~\hfill~71.50~\hfill~ & ~\hfill~56.39~\hfill~ & ~\hfill~66.81~\hfill~ & ~\hfill~67.92~\hfill~ & ~\hfill~73.58~\hfill~ & ~\hfill~48.37~\hfill~ \\
\multicolumn{1}{c|}{ELECTRA}  & ~\hfill~75.30~\hfill~ & ~\hfill~54.83~\hfill~ & ~\hfill~68.80~\hfill~ & ~\hfill~67.02~\hfill~ & ~\hfill~73.16~\hfill~ & ~\hfill~50.98~\hfill~ \\
\multicolumn{1}{c|}{OpenAI-GPT}  & ~\hfill~70.87~\hfill~ & ~\hfill~47.44~\hfill~ & ~\hfill~64.10~\hfill~ & ~\hfill~64.15~\hfill~ & ~\hfill~68.06~\hfill~ & ~\hfill~52.29~\hfill~ \\
\multicolumn{1}{c|}{Tranformer-XL}  & ~\hfill~61.95~\hfill~ & ~\hfill~42.67~\hfill~ & ~\hfill~59.54~\hfill~ & ~\hfill~60.89~\hfill~ & ~\hfill~64.36~\hfill~ & ~\hfill~47.06~\hfill~ \\
\multicolumn{1}{c|}{CTRL}  & ~\hfill~62.29~\hfill~ & ~\hfill~41.11~\hfill~ & ~\hfill~56.84~\hfill~ & ~\hfill~57.01~\hfill~ & ~\hfill~60.55~\hfill~ & ~\hfill~39.87~\hfill~ \\
\multicolumn{1}{c|}{OpenAI-GPT2}  & ~\hfill~55.79~\hfill~ & ~\hfill~42.72~\hfill~ & ~\hfill~60.11~\hfill~ & ~\hfill~51.55~\hfill~ & ~\hfill~63.42~\hfill~ & ~\hfill~40.52~\hfill~ \\

\bottomrule
\end{tabular}%
}

\end{table*}

\section{Conclusion}

In this paper, we evaluated the contextualisation power of nine pre-trained Transformer Models on a WSD task. We presented a comparative study on each model's power to capture polysemy in the embeddings they generate. To accomplish this, we used a kNN based approach to WSD proposed by \cite{Wiedemann2019DoesBM} and proposed two improvements in their method that also accompanied us in establishing a new state-of-the-art on WSD Lexical Sample Task of SensEval-2 and SensEval-3. We concluded our study by stating that BERT, DistilBERT, and ALBERT models prove out to be most effective on the WSD task solely based on text encodings they provide. We found these models to possess an extraordinary potential to identify a word's different senses compared to all the other models. 

As future work, we plan to make use of POS information as well to classify an ambiguous word. We firmly believe that incorporating POS information in WSD could be very useful and further increase the performance of these models. In addition to this, we also believe that fine-tuning these models could be a potential area of focus. In our experiments, we leveraged the pre-trained models as provided by the authors, and a bit of fine-tuning could be beneficial.

\bibliographystyle{splncs04}

\bibliography{references.bib}

\end{document}